\documentclass[letterpaper]{article} 
\usepackage{aaai24}  
\usepackage{times}  
\usepackage{helvet}  
\usepackage{courier}  
\usepackage[hyphens]{url}  
\usepackage{graphicx} 
\usepackage{natbib}  
\usepackage{caption} 
\usepackage{algorithm}
\usepackage{algorithmic}

\usepackage{newfloat}
\usepackage{listings}
\DeclareCaptionStyle{ruled}{labelfont=normalfont,labelsep=colon,strut=off} 
\floatstyle{ruled}
\newfloat{listing}{tb}{lst}{}
\floatname{listing}{Listing}
\usepackage{amsfonts,amssymb}
\usepackage{threeparttable}
\usepackage{booktabs}
\usepackage{diagbox}
\usepackage{colortbl}

\definecolor{orange}{RGB}{255,140,0}
\definecolor{myblue}{RGB}{30,144,255}
\definecolor{mypurple}{RGB}{138,44,226}

\usepackage{multirow}
\newcommand{\tabincell}[2]{\begin{tabular}{@{}#1@{}}#2\end{tabular}}
\usepackage{amsmath,bm}
\usepackage{dsfont}
\usepackage{color} 
\usepackage[switch]{lineno}  %

\usepackage{xspace}

\title{Leveraging Imagery Data with Spatial Point Prior for\\ Weakly Semi-supervised 3D Object Detection}

\author{
    Hongzhi Gao\textsuperscript{\rm 1}, 
    Zheng Chen\textsuperscript{\rm 1}, 
    Zehui Chen\textsuperscript{\rm 1}\footnote{Project Leader}, 
    Lin Chen\textsuperscript{\rm 1}, \\
    Jiaming Liu\textsuperscript{\rm 2}, 
    Shanghang Zhang\textsuperscript{\rm 2},
    Feng Zhao\textsuperscript{\rm 1}\footnote{Corresponding Author} \\
}
\affiliations{
    \textsuperscript{\rm 1} University of Science and Technology of China\\
    \textsuperscript{\rm 2} Peking University \\
    \{hongzhigao, chenzheng, lovesnow, chlin\}@mail.ustc.edu.cn, \\
    liujiaming@bupt.cn, shanghang@pku.edu.cn, 
    fzhao956@ustc.edu.cn

}

\usepackage{bibentry}

\begin{document}

\maketitle

\begin{abstract}
    Training high-accuracy 3D detectors necessitates massive labeled 3D annotations with 7 degree-of-freedom, which is laborious and time-consuming. 
    Therefore, the form of point annotations is proposed to offer significant prospects for practical applications in 3D detection, which is not only more accessible and less expensive but also provides strong spatial information for object localization.
    In this paper, we empirically discover that it is non-trivial to merely adapt Point-DETR to its 3D form, encountering two main bottlenecks: 1) it fails to encode strong 3D prior into the model, and 2) it generates low-quality pseudo labels in distant regions due to the extreme sparsity of LiDAR points. 
    To overcome these challenges, we introduce Point-DETR3D, a teacher-student framework for weakly semi-supervised 3D detection, designed to fully capitalize on point-wise supervision within a constrained instance-wise annotation budget.
    Different from Point-DETR which encodes 3D positional information solely through a point encoder, we propose an explicit positional query initialization strategy to enhance the positional prior. 
    Considering the low quality of pseudo labels at distant regions produced by the teacher model, 
    we enhance the detector's perception by incorporating dense imagery data through a novel Cross-Modal Deformable RoI Fusion (D-RoI).
    Moreover, an innovative point-guided self-supervised learning technique is proposed to allow for fully exploiting point priors, even in student models.
    Extensive experiments on representative nuScenes dataset demonstrate our Point-DETR3D obtains significant improvements compared to previous works. Notably, with only 5\% of labeled data, Point-DETR3D achieves over 90\% performance of its fully supervised counterpart.
\end{abstract}

\section{Introduction}
\label{sec:intro}

3D object detection is one of the fundamental tasks in autonomous driving perception.
In recent years, contemporary 3D object detectors \cite{lang2019pointpillars,wang2022detr3d,chen2022graph,wang2021object,chen2022autoalign,chen2022autoalignv2} have made significant strides. 
Current advanced detectors typically necessitate training on myriad scenes with precise 3D annotations, delineating the 3D objects' locations, dimensions, and orientations using 7 degree-of-freedom (DoF).
However, the manual annotation process for 3D labels is time-consuming and expensive \cite{su2012crowdsourcing}, emphasizing the need to diminish dependence on massive 3D box annotations.

\begin{figure}[!t]
\centering
\includegraphics[width=1\linewidth]{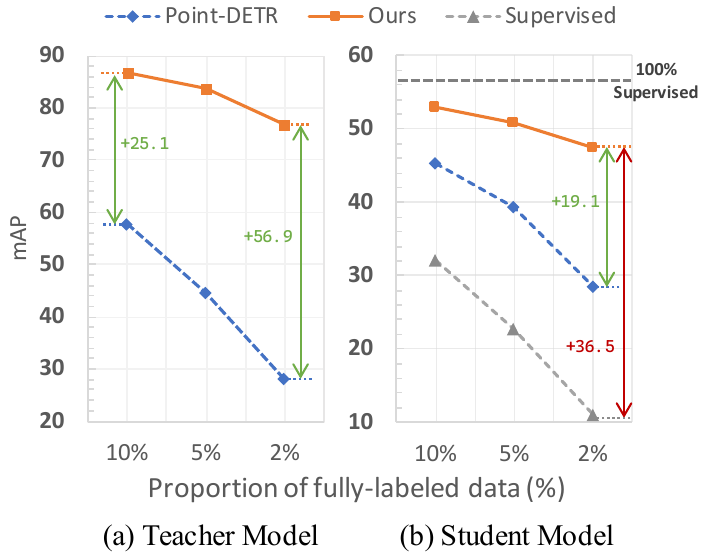}
\caption{Comparison of mAP on teacher and student models. (a) Our teacher model outperforms a large margin over the Point-DETR baseline. (b) Our student model which only uses 10\% fully-labeled data achieves comparable performance with the 100\% supervised paradigm on the CenterPoint baseline.
}
\vspace{-0.5cm}
\label{fig:1}
\end{figure}

\begin{figure*}[h]
\centering
\includegraphics[width=1.  \textwidth]{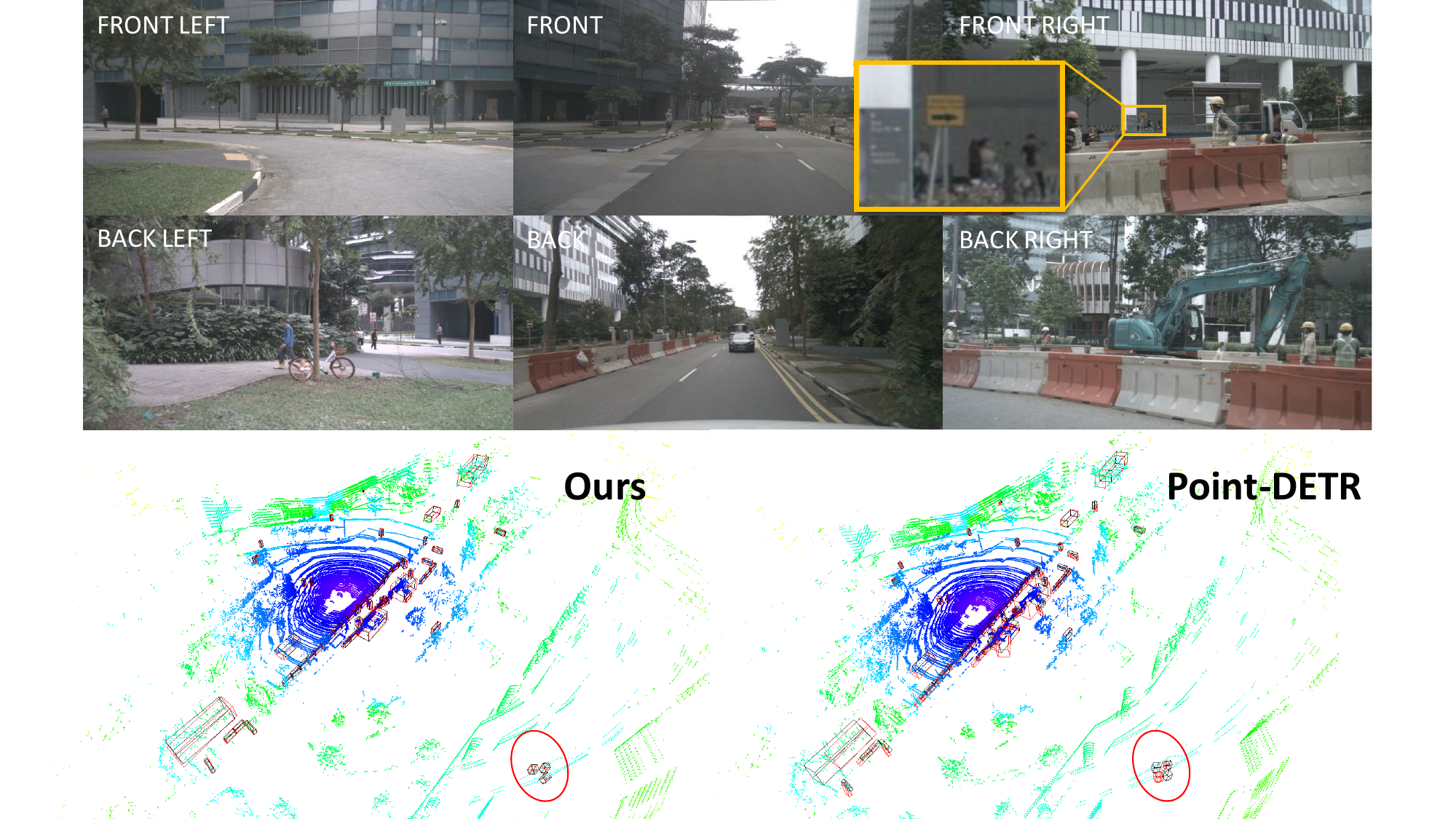}
\caption{
    Comparison of Pseudo Label Quality for Point-DETR3D (Ours) and Point-DETR. For the point cloud visualization, we show the prediction and ground truths in \textcolor{red}{red} and black, respectively. 
    We can observe that Point-DETR3D can accurately detect far-distant objects, but some of these objects are regressed poorly in Point-DETR, as indicated by the red circle in the figure. 
    Best viewed with color and zoom-in.
}
\label{fig.framework}
\end{figure*}

To mitigate manual annotation efforts, previous studies 
have proposed weakly-supervised \cite{meng2020weakly,peng2022weakm3d,xu2022back} and semi-supervised \cite{zhao2020sess,yin2022semi,park2022detmatch} methods for 3D object detection. These strategies either leverage incomplete annotations or utilize a small subset of meticulously annotated scenes combined with a more extensive collection of unlabeled scenes for training. Nevertheless, they yield limited improvements in detection performance.
Weakly Semi-Supervised Learning (WSSL) \cite{li2018weaklysemi,chen2021points,zhang2022group} bridges the gap between these approaches, offering a balance between labeling costs and model efficacy by annotating large-scale scenes with only one point per instance and a few scenes with 3D box annotations.

However, weakly semi-supervised 3D object detection using points remains an under-explored area.  
Modern WSSL frameworks primarily focus on 2D object detection \cite{chen2021points,zhang2022group} and 2D segmentation \cite{kim2023devil,papandreou2015weaklysemi,li2018weaklysemi,lee2019ficklenet}. Despite the impressive performance in these tasks, we empirically find that applying them to 3D object detection is non-trivial, mainly due to overlooking the inherent characteristics of LiDAR points. In this paper, we start the point by adapting the competitive WSSL 2D-based Point-DETR \cite{chen2021points} to a 3D-based WSSL detector, and identify two core impediments for the final detection performance. 
On the one hand, even though Point-DETR explicitly encodes the coordinates and category information through the proposed Point Encoder, it still struggles with limited positional priors, making instance identification challenging. 
On the other hand, it generates extremely low-quality pseudo labels in distant regions, likely due to the extreme sparsity of LiDAR points (\textit{i.e.}, with only a few LiDAR points, accurate shape, and position annotation become unfeasible, even for skilled human experts). 

To address these challenges, we introduce the so-called Point-DETR3D, a weakly semi-supervised 3D object detection framework designed to fully capitalize on point-level annotations within a constrained instance-wise label budget.
Firstly, we propose an explicit positional query initialization strategy, rather than merely encoding 3D point coordinates through a learnable point encoder. This approach seamlessly integrates absolute 3D positions with object queries, initializing them directly from point annotations, and significantly accelerates model convergence (\textit{i.e.}, our model can generate high-quality 3D bounding boxes with only one transformer decoder).
Secondly, considering the characteristics of LiDAR points, we leverage the dense imagery data to assist the pseudo-label generation in distant regions. Specifically, we propose a novel cross-modal feature interaction module, namely Deformable RoI Fusion Module, to dynamically aggregate 2D information for accurate boundary localization. 
Furthermore, we introduce a general point-guided self-supervised learning paradigm for the student model, which makes full use of point annotations in a parameter-free manner, substantially mitigating pseudo-label noise and enhancing representation robustness.

In summary, the contributions are three-fold:
\begin{itemize}
\item We identify the main bottlenecks of the current WSSL framework in 3D detection, which mainly attributes to the insufficient usage of 3D point priors and the sparsity characteristic of LiDAR inputs. 
\item Based on the above observations, we propose a new WSSL framework for 3D object detection, namely Point-DETR3D, which consists of a point-centric teacher model and a self-motivated student model. 
\item Extensive experiments on the representative nuScenes benchmark demonstrate the effectiveness of our proposed detector. Notably, with only 5\% of labeled data, Point-DETR3D achieves over 90\% performance of its fully supervised counterpart.
\end{itemize}

\begin{figure*}[h]
\centering
\includegraphics[width=1.  \textwidth]{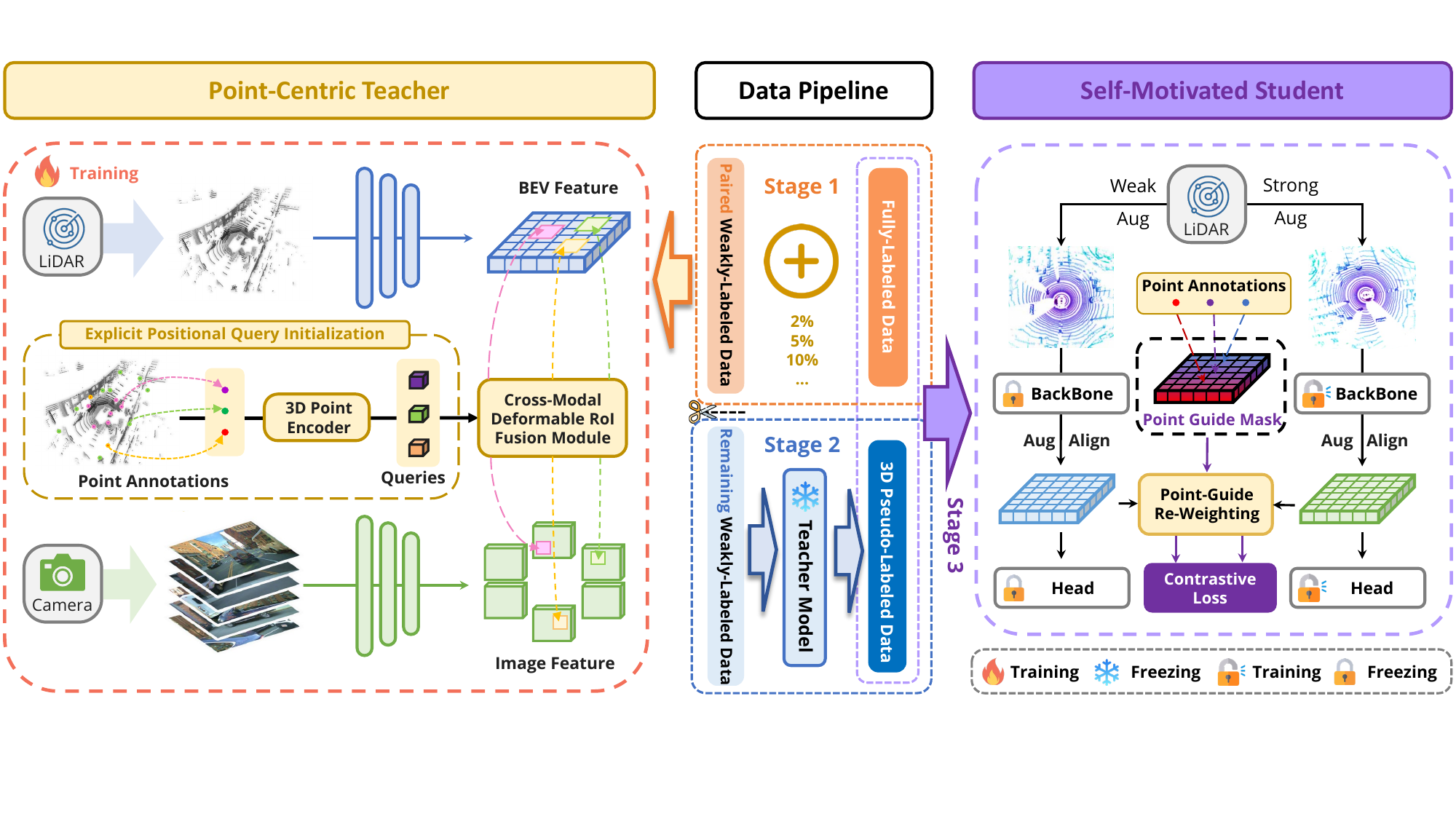}
\vspace{-2.1cm}
\caption{
    The overall framework of Point-DETR3D. The stage 1 is the training stage of the teacher model which utilizes limited fully-labeled data and paired weakly-labeled data as the training set. The stage 2 represents the pseudo-label generation stage of the teacher model, which transforms the remaining weakly-labeled data into 3D pseudo boxes. The stage 3 is the self-supervised training process of the student model, which is trained with the full set of both labeled and pseudo annotations.
}
\label{fig.framework}
\end{figure*}

\section{Related Work}
\label{sec:related}

\subsubsection{3D Object Detection.} 
Existing 3D object detectors can be broadly categorized as uni-modal and cross-modal approaches. Uni-modal methods utilize point clouds \cite{zhou2018voxelnet,lang2019pointpillars} or images \cite{wang2022detr3d,chen2022graph} only as inputs. VoxelNet \cite{zhou2018voxelnet} and PointPillar \cite{lang2019pointpillars} processes the point features at voxel or BEV space, respectively. CenterPoint \cite{yin2021center} revisits the label assignment in 3D detection and proposes a general center-based assignment strategy. Object-DGCNN \cite{wang2021object} and SST \cite{2021Embracing} prove that transformer-based detectors can also yield competitive results.
Recently, there has been increasing attention on multi-modal data to further boost the 3D detection performance. TransFusion \cite{2022transfusion} incorporates imagery information through a transformer-style fusion module. BEVFusion \cite{liang2022bevfusion} directly concatenates the different modalities by unifying them into the same BEV space.
DeepInteraction \cite{yang2022deepinteraction} introduces a modality interaction strategy that leverages representational and predictive interaction to prevent the loss of unique perception strengths.
SparseFusion \cite{xie2023sparsefusion} proposes instance-level sparse feature fusion and cross-modality information transfer to take advantage of the strengths of each modality while mitigating their weaknesses.
Despite being effective, how to utilize imagery information with point prior is still under-explored.




\subsubsection{Semi-Supervised 3D Object Detection.} 
Although supervised 3D object detection methods yield promising results, the heavy reliance on massive accurate annotations makes it hard to be deployed in real-world scenarios. Semi-supervised 3D detection presents a potential solution to alleviate the labeling cost issue, which only requires a small fraction of labeled data.
Based on the Mean Teacher \cite{tarvainen2017mean} paradigm, SESS \cite{zhao2020sess} applies a consistency loss between teacher predictions and student predictions. DetMatch \cite{park2022detmatch} introduces a flexible framework for joint semi-supervised learning on 2D and 3D modalities, generating more robust pseudo-labels. The quality of the pseudo-labels is further enhanced by 3DIoUMatch \cite{wang20213dioumatch}, which employs a dedicated filter approach. ProficientTeacher \cite{yin2022semi} goes a step further by proposing an STE module to generate sufficient seed boxes, and then the CBV module is used to select high-quality predictions.

\subsubsection{Weakly Semi-Supervised Object Detection.} 
Previous works \cite{bilen2016weakly,tang2017multiple,yang2019towards} commonly utilize image-level annotations as weak supervision. However, the absence of location information greatly hinders the model's performance. 
Recent works propose to leverage point annotations as weak supervision signals. 
WS3D \cite{meng2020weakly} devises a two-stage framework that achieves competitive performance by utilizing only a small set of weakly annotated scenes, along with a few precisely labeled object instances.
Point-Teaching \cite{ge2023point} applies a Hungarian-based point-matching method to obtain pseudo labels, which are then utilized in conjunction with multiple instance learning (MIL) approaches to supervise the object detector. To leverage point annotations, Point-DETR \cite{chen2021points} introduces a transformer-based point-to-box regressor that transforms point annotations into pseudo boxes so that a student object detector can be trained. Group R-CNN \cite{zhang2022group} designs a RPN-based point-to-box projector, which enhances RPN recall by establishing an Instance Group and further improves accuracy through the adoption of a new assignment strategy. However, most of works are applicable only to 2D target detection. In the realm of 3D detection, there remains a need to further investigate weakly semi-supervised approaches.


\section{Point-DETR3D}
\label{sec:method}

In this section, we first review the preliminaries of weakly semi-supervised 3D object detection with point annotations (WSS3D-P). After that, we introduce Point-DETR3D, an effective teacher-student-based framework for WSS3D-P. It consists of two main components: the Point-Centric Teacher  and the Self-Motivated Student. The overall framework is shown in Figure \ref{fig.framework}.
\subsection{Preliminaries}
\subsubsection{3D Point Annotations.} 
Each 3D point annotation includes an object point and its category label. 
We form each object as $(x,y,z,c)$, where $(x, y, z)$ and $c$ represent point location in 3D space and object category, respectively. 
Note that the object point annotations are randomly sampled within the 3D box following a normal distribution prior. Such an approach enables us to dramatically alleviate the massive burden of precisely annotating objects in 3D tasks. 

\subsubsection{WSS3D-P Pipeline.} 
WSS3D-P is a challenging task due to point annotations providing limited information, making it difficult to precisely predict 3D bounding boxes in a scene. 
It typically utilizes a small portion of well-labeled scenes and a large number of weakly point-level annotations as training data. 
Previous works adapt the teacher-student training paradigm as the default training pipeline, which leads to significant progress in semi-supervised learning \cite{wang20213dioumatch,yin2022semi}. 
The steps are summarized as follows:

\textbf{(1)} {\bf Train a point-to-box teacher model} with a small portion of paired point- and fully-annotated data.

\textbf{(2)} {\bf Generate pseudo labels for weakly-labeled scenes} using the trained teacher model.

\textbf{(3)} {\bf Train a student model} with the combination of fully-labeled scenes and the remaining pseudo-labeled scenes. 

\subsection{Point-Centric Teacher}
\label{sec.teacher}
The Point-Centric Teacher is designed to generate precise 3D pseudo boxes by fully capitalizing on the potential of weak point annotations. This is achieved through two core components: the Explicit Positional Query Initialization strategy and the Cross-Modal Deformable RoI Fusion module, which together seamlessly harness strong positional priors and dense visual features for precise localization.

\subsubsection{Explicit Positional Query Initialization.}

While Point-DETR \cite{chen2021points} utilizes point annotations by encoding both coordinates and category information through a learnable projector, such an approach yields minimal improvements when extended to the 3D domain. This limitation may be attributed to the significant increase in candidate space when lifting from 2D to 3D. 
To address this issue, we opt for an explicit positional query initialization strategy rather than merely embedding prior information implicitly via a 3D point encoder. 
Specifically, we enable an explicit bind between the absolute 3D position and each object query. Thanks to the generality of Object-DGCNN \cite{wang2021object}, this bind can be easily achieved through the \textit{reference point}. To this end, we can generate $n$ object queries, and each reference point of them is initialized as the $n$ GT point annotations. Since there is a direct correlation between each object query and GT, we replace the originally complex bipartite matching with a straightforward one-to-one match based on such initialization, which significantly reduces the training instability observed in \cite{li2022dn}.

\subsubsection{Point-Centric Deformable RoI Fusion.}

\begin{figure}[!t]
\centering
\includegraphics[width=0.99\linewidth]{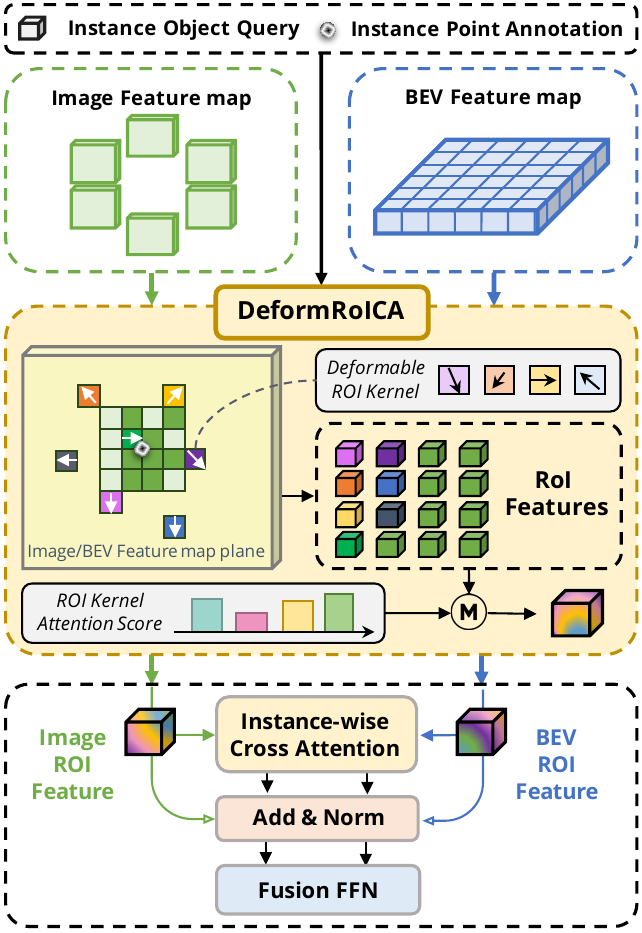}
\caption{
Illustration of the Point-Centric Deformable RoI Cross-Modal Fusion Module.
First, it projects the instance point annotations onto the image and Bird's Eye View (BEV) plane, generating RoI-wise grid points. These grid points function as reference points and can be deformed through learnable coordinate offsets. Subsequently, each object query interacts with the corresponding RoI features to assimilate both LiDAR and imagery information.
Finally, an instance-wise cross-attention module is applied to aggregate instance features from both modalities. 
}
\label{fig.fusion_module}
\end{figure}

Despite being informed with strong 3D points prior, we empirically observe that our teacher model produces subpar results in distant regions. 
In fact, even a skilled human expert would struggle to determine the exact boundaries of target instances with only a few points.
To make up for this information deficit, we introduce dense imagery data as a valuable reference.
Previous works \cite{yang2022deepinteraction,chen2022autoalign,xie2023sparsefusion} have demonstrated remarkable performance gains by combining 2D and 3D data in 3D object detection. Yet, most works mainly focus on general cross-modal fusion at a voxel/point level, while effective instance-level fusion with 3D priors has rarely been explored. In this work, we present a novel Point-Centric Deformable RoI Cross-Modal Fusion operation (see Figure \ref{fig.fusion_module}), which seamlessly aggregates instance-level features guided by the 3D prior through RoI-wise point sampling and dynamically decides the imagery regions for reference, especially in distant regions.

Formally, given a 3D instance point annotation $p^{3d}_i=(x_i, y_i, z_i)$, we first project it onto different image views using the projection matrix of the respective cameras. The rotation and translation matrices from the LiDAR coordinate to the camera coordinate of the $j^{th}$ view are represented by $R_j \in \mathbb{R}^{3 \times 3}$ and $t_j \in \mathbb{R}^{3}$, respectively, and the intrinsic parameter of the $j^{th}$ camera is denoted as $K_j \in \mathbb{R}^{3 \times 3}$.
The point projection from LiDAR to the camera can be written as:

\begin{equation}
    p^{img}_{i}(u_i,v_i) = K_j(R_j p^{3d}_i + t_j),
\end{equation}
where $p^{img}_{i}=(u_i,v_i)$ is image coordinates of point annotations.

To obtain the RoI reference points $p^{RoI}_{ik}$, we generate a square grid centered around $p^{img}_{i}$. Here, $k$ refers to the index of the RoI reference points in the RoI region $(k \in {1,..., K^2}, e.g., K=4)$, while $K$ serves as the square grid size. 
Unlike in previous works\cite{yang2022deepinteraction,chen2022autoalign}, we do not simply sample RoI features using reference points. Instead, we introduce a learnable offset $\Delta_{ik}$ to each RoI reference point, which dynamically determines the most informative neighboring features for aggregation and is not limited to fixed RoI regions.
The process of deformable RoI cross-attention (DRoICA) can be formulated as:

\begin{equation}
\centering
\begin{aligned}
    \operatorname{DRoICA}(Q_i, F)= \hspace{130pt} \\
    \sum_{k=1}^{K}  \operatorname{MSDeformAttn}(Q_i, (p^{RoI}_{mlik} + \Delta_{mlik}), \{F_l\}^{L}_{l=1}),
\end{aligned}
\end{equation}

where $m$ indexes the attention head and $l$ indexes the input feature level. $Q_i$ denotes the instance query and $F$ is the corresponding multi-level image features with the RoI center point $p^{img}_{i}$. 
For the BEV feature map, we follow the same process to generate RoI grids for each instance query in the 2D plane. We then apply deformable RoI cross-attention to obtain the instance feature $F^{I}_{i}$ and $F^{L}_{i}$ from camera and LiDAR features.
To maintain the spatial localization information from the LiDAR RoI feature and the textural information from the image RoI feature, the corresponding features from the two modalities interact at the instance level through a cross-attention module proposed in \cite{vaswani2017attention}.



\subsection{Self-Motivated Student}
\label{sec.student}
In the original Point-DETR, the student model is solely trained on a combination of fully labeled data and pseudo-labeled data generated by the teacher model. 
However, this approach overlooks the potential benefits that the student model could derive from point annotations during training. 
In order to maximize the utilization of prior information of points and enhance the final performance of the model, we propose a simple and parameter-free self-supervised approach called Point-Centric Feature-Invariant Learning. It particularly focuses on mitigating the influence of label noises and strengthening the robustness of the model's representations.

Drawing inspiration from \cite{chen2021exploring, chen2020simple}, we employ the standard contrastive learning training paradigm as our foundational framework (see Figure \ref{fig.framework}).
For a given input $x$, it is fed into two randomly augmented pipelines corresponding to weak and strong augmentations, producing two distinct inputs $x_{1}$ and $x_{2}$, respectively. Both of these views are processed by a feature extractor to derive BEV features. Subsequently, we reverse the BEV features given the augmented pipelines to generate the corresponding paired features for consistency regularization. Previous CL works\cite{liang2021expcontrast,chen2021exploring} directly force the student to mimic the feature map with equal supervision:
\begin{equation}
	L_{feat} = \frac{1}{H \times W} \sum_i^H\sum_j^W ||F_{ij}^{2D} - F_{ij}^{2D}||_2,
\end{equation}
where $H, W$ denotes the width and height of the image feature map, and $||\cdot||_2$ is the $L_2$ norm. 
However, considering the characteristics of LiDAR points, there is a large portion of empty space in the BEV plane. Conducting supervision on these \textit{NULL} representations makes no contribution to the model learning or even deteriorates the final performance. Therefore, we leverage the point annotations as foreground regional prompts to guide the learning positions.
Concretely, we generate a Gaussian distribution for each point annotation $(x_i, y_i)$ in the BEV space following \cite{yin2021center}:
\begin{equation}
    m_{i,x,y} = \text{exp}(-\frac{(x_i - \hat{x}_i)^2 + (y_i - \hat{y}_i)^2}{2 \sigma_i^2}),
\end{equation}
where $\sigma_i$ is a constant (default value of 2). 
Since the feature maps are class-agnostic, we combine all $m_{i,x,y}$ into one single mask $M$. For overlapping regions among different $m_{i,x,y}$ at the same position, we simply take the maximum value of them. 
After generating the mask $M$, we use it to guide the student in mimicking the feature for dense feature contrastive learning:
\begin{equation}
\centering
\begin{aligned}
    L_{feat} = \frac{1}{H \times W \times \sum max(M_{i,j})} \hspace{40pt} \\
  \sum_i^H\sum_j^W max(M_{ij}) ||F_{ij}^{3D} - F_{ij}^{3D}||_2 \\
\end{aligned}
\end{equation}
Such a point-guided re-weighting strategy enables the model to concentrate on the foreground regions from the teacher while avoiding the useless imitation of empty 3D features in the excessive background regions.

\section{Experiments}
\label{sec:experiment}

\subsection{Dataset and Evaluation Metrics}

\textbf{Dataset.}
We conduct the experiments on the nuScenes dataset\cite{nuscenes}, a well-established dataset in the domain of 3D object detection. It consists of 700 training scenes, 150 validation scenes, and 150 testing scenes. For each scene, it provides point clouds collected from a 32-beam LiDAR system and images captured by 6 surrounding cameras with a resolution of 1600 × 900. For the 3D object detection task, 1.4M objects are annotated with 3D bounding boxes and classified into 10 categories. \\
\textbf{Evaluation Metrics.}
We report \textit{Static Properties nuScenes Detection Score} (SPNDS) and mean Average Precision (mAP) as our evaluation metrics. 
Specifically, SPNDS is derived from standard NDS by removing the velocity-related measurements including mean Average Velocity Error (mAVE) and mean Average Attribute Error (mAAE), since we only focus on the static attributes of 3D bounding boxes. 
Formally, SPNDS is a weighted sum of mean Average Precision (mAP) and the aforementioned True Positive (TP), which is defined as: 
\begin{equation}
SPNDS = \frac{1}{8} \left[ 5 \times mAP + \sum_{\text{mTP} \in TP}{(1 - \text{min}(1, \text{mTP}))} \right]
\end{equation}
\subsection{Implementation Details}

Our codebase is built on MMDetection3D toolkit. All models are trained on an 8 NVIDIA A100 GPU machine. 
Following prior work\cite{chen2021points}, we sample 2{\%}, 5{\%}, and 10{\%} of training scenes as the fully labeled set, and the rest scenes are set as weakly labeled. We employ a sequential sampling approach to avoid data
leakage.
To maintain consistency with Point-DETR, we select a transformer-based 3D detector Object-DGCNN\cite{wang2021object} as our default teacher model. 
Note that any 3D detector can be set as the student model only if it provides BEV features. Without loss of generality, we choose CenterPoint \cite{yin2021center} with both pillar-based and voxel-based settings.
\subsection{Main Results}
In this section, we report the experimental results on teacher and student models given different ratios of fully labeled sets. For a fair comparison, we carefully modify the Point-DETR to adapt it to the 3D domain, which serves as a competitive baseline.


\begin{table}[]
\begin{tabular}{@{}ccccc@{}}
\toprule
Backbone & Ratio & Paradigm & SPNDS ↑ & mAP ↑ \\ \midrule
\multicolumn{1}{c|}{\multirow{6}{*}{Pillar-based}} & \multicolumn{1}{c|}{\multirow{2}{*}{2\%}} & \multicolumn{1}{c|}{Point-DETR} & 23.87 & 23.55 \\
\multicolumn{1}{c|}{} & \multicolumn{1}{c|}{} & \multicolumn{1}{c|}{Ours} & \textbf{66.03} & \textbf{80.47} \\ \cmidrule(l){2-5} 
\multicolumn{1}{c|}{} & \multicolumn{1}{c|}{\multirow{2}{*}{5\%}} & \multicolumn{1}{c|}{Point-DETR} & 35.28 & 39.89 \\
\multicolumn{1}{c|}{} & \multicolumn{1}{c|}{} & \multicolumn{1}{c|}{Ours} & \textbf{69.18} & \textbf{84.09} \\ \cmidrule(l){2-5} 
\multicolumn{1}{c|}{} & \multicolumn{1}{c|}{\multirow{2}{*}{10\%}} & \multicolumn{1}{c|}{Point-DETR} & 51.33 & 61.41 \\
\multicolumn{1}{c|}{} & \multicolumn{1}{c|}{} & \multicolumn{1}{c|}{Ours} & \textbf{73.19} & \textbf{86.50} \\ \midrule
\multicolumn{1}{c|}{\multirow{6}{*}{Voxel-based}} & \multicolumn{1}{c|}{\multirow{2}{*}{2\%}} & \multicolumn{1}{c|}{Point-DETR} & 27.49 & 28.19 \\
\multicolumn{1}{c|}{} & \multicolumn{1}{c|}{} & \multicolumn{1}{c|}{Ours} & \textbf{63.98} & \textbf{76.82} \\ \cmidrule(l){2-5} 
\multicolumn{1}{c|}{} & \multicolumn{1}{c|}{\multirow{2}{*}{5\%}} & \multicolumn{1}{c|}{Point-DETR} & 39.28 & 44.56 \\
\multicolumn{1}{c|}{} & \multicolumn{1}{c|}{} & \multicolumn{1}{c|}{Ours} & \textbf{70.28} & \textbf{83.63} \\ \cmidrule(l){2-5} 
\multicolumn{1}{c|}{} & \multicolumn{1}{c|}{\multirow{2}{*}{10\%}} & \multicolumn{1}{c|}{Point-DETR} & 49.35 & 57.58 \\
\multicolumn{1}{c|}{} & \multicolumn{1}{c|}{} & \multicolumn{1}{c|}{Ours} & \textbf{76.46} & \textbf{87.46} \\ \bottomrule
\end{tabular}
\caption{Comparison of different teacher models on the validation set with varying ratios of labeled data. Note that point annotations are used in Point-DETR and ours(Point-DETR3D) during training and validation. ResNet-50 is the default image backbone in Point-DETR3D.}
\label{tab:teacher}
\end{table}

\subsubsection{Teacher Models Performance.}

Table \ref{tab:teacher} compares the results of our teacher model with Point-DETR under varying ratios of fully labeled data. Our proposed Point-DETR3D outperforms Point-DETR by more than 20.0 SPNDS and 30.0 mAP across all data configurations. Such significant improvements highlight the efficacy of our explicit positional query initialization coupled with the incorporation of imagery data. 
Interestingly, the pillar-based teacher yields similar performance to the voxel-based one, suggesting that our models are nearing the upper bound of detection accuracy when only using point annotations.

\subsubsection{Student Models Performance.}

We evaluate the benefits of different training paradigms by comparing the student model CenterPoint \cite{yin2021center}, which is trained on the pseudo labels generated from the corresponding teacher (pillar/voxel-based).
The results are shown in Table \ref{tab:student} and Table \ref{tab:focalformer3d}. Both Point-DETR and Point-DETR3D markedly outperform the vanilla baseline, demonstrating the merit of weakly semi-supervised learning with point annotations. 
Moreover, Point-DETR3D leads the original Point-DETR by 5.5 $\sim$ 19.9 SPNDS under various labeled data conditions, substantiating the improved pseudo-label quality and the advantage of the point-centric feature-invariant learning module.
It's worth noting that with only 5\% fully labeled samples, our student model achieves almost 90\% performance of a fully supervised 3D detector, which demonstrates that our Point-DETR3D is able to dramatically reduce the labeling efforts on 3D annotations.

\begin{table}[!t]
\centering
\begin{tabular}{@{}ccccc@{}}
\toprule
3D Detector & Ratio & Paradigm & SPNDS & mAP \\ \midrule
\multicolumn{1}{c|}{} & \multicolumn{1}{c|}{100\%} & \multicolumn{1}{c|}{-} & 56.28 & 48.90 \\ \cmidrule(l){2-5} 
\multicolumn{1}{c|}{\multirow{9}{*}{\tabincell{c}{CenterPoint\\-Pillar}}} & \multicolumn{1}{c|}{\multirow{3}{*}{2\%}} & \multicolumn{1}{c|}{-} & 16.97 & 5.52 \\
\multicolumn{1}{c|}{} & \multicolumn{1}{c|}{} & \multicolumn{1}{c|}{Point-DETR} & 24.40 & 21.02 \\
\multicolumn{1}{c|}{} & \multicolumn{1}{c|}{} & \multicolumn{1}{c|}{Ours} & \textbf{40.09} & \textbf{39.49} \\ \cmidrule(l){2-5} 
\multicolumn{1}{c|}{} & \multicolumn{1}{c|}{\multirow{3}{*}{5\%}} & \multicolumn{1}{c|}{-} & 23.85 & 13.34 \\
\multicolumn{1}{c|}{} & \multicolumn{1}{c|}{} & \multicolumn{1}{c|}{Point-DETR} & 35.14 & 30.75 \\
\multicolumn{1}{c|}{} & \multicolumn{1}{c|}{} & \multicolumn{1}{c|}{Ours} & \textbf{45.81} & \textbf{41.56} \\ \cmidrule(l){2-5} 
\multicolumn{1}{c|}{} & \multicolumn{1}{c|}{\multirow{3}{*}{10\%}} & \multicolumn{1}{c|}{-} & 31.29 & 20.44 \\
\multicolumn{1}{c|}{} & \multicolumn{1}{c|}{} & \multicolumn{1}{c|}{Point-DETR} & 45.07 & 39.00 \\
\multicolumn{1}{c|}{} & \multicolumn{1}{c|}{} & \multicolumn{1}{c|}{Ours} & \textbf{48.96} & \textbf{43.86} \\ \midrule
\multicolumn{1}{c|}{} & \multicolumn{1}{c|}{100\%} & \multicolumn{1}{c|}{-} & 61.80 & 56.37 \\ \cmidrule(l){2-5} 
\multicolumn{1}{c|}{\multirow{9}{*}{\tabincell{c}{CenterPoint\\-Voxel}}} & \multicolumn{1}{c|}{\multirow{3}{*}{2\%}} & \multicolumn{1}{c|}{-} & 20.82 & 11.01 \\
\multicolumn{1}{c|}{} & \multicolumn{1}{c|}{} & \multicolumn{1}{c|}{Point-DETR} & 29.14 & 28.40 \\
\multicolumn{1}{c|}{} & \multicolumn{1}{c|}{} & \multicolumn{1}{c|}{Ours} & \textbf{48.21} & \textbf{47.49} \\ \cmidrule(l){2-5} 
\multicolumn{1}{c|}{} & \multicolumn{1}{c|}{\multirow{3}{*}{5\%}} & \multicolumn{1}{c|}{-} & 33.21 & 22.82 \\
\multicolumn{1}{c|}{} & \multicolumn{1}{c|}{} & \multicolumn{1}{c|}{Point-DETR} & 42.55 & 39.26 \\
\multicolumn{1}{c|}{} & \multicolumn{1}{c|}{} & \multicolumn{1}{c|}{Ours} & \textbf{53.99} & \textbf{50.80} \\ \cmidrule(l){2-5} 
\multicolumn{1}{c|}{} & \multicolumn{1}{c|}{\multirow{3}{*}{10\%}} & \multicolumn{1}{c|}{-} & 42.18 & 32.07 \\
\multicolumn{1}{c|}{} & \multicolumn{1}{c|}{} & \multicolumn{1}{c|}{Point-DETR} & 50.58 & 45.30 \\
\multicolumn{1}{c|}{} & \multicolumn{1}{c|}{} & \multicolumn{1}{c|}{Ours} & \textbf{57.81} & \textbf{53.03} \\ \bottomrule
\end{tabular}
\caption{Comparison of different student models which are trained with the combined of labeled data and pseudo labels generated by difference teacher models. ``-'' means the model is trained on the full labeled data only.}
\label{tab:student}
\end{table}

\begin{table}[]
\centering
\vspace{-0.2cm}
\begin{tabular}{@{}c|c|c|c@{}}
\toprule
Paradigm & 2\% & 5\% & 10\%  \\ \midrule
-           & 21.71/10.29 & 36.90/25.61 & 46.60/35.96  \\
Point-DETR  & 30.55/30.55 & 43.88/42.81 & 54.56/51.26     \\
Ours$^\dagger$   & \textbf{51.04/54.53} & \textbf{56.39/57.15} & \textbf{60.93/58.99}     \\ \bottomrule
\end{tabular}
\caption{Additional experiments of SPNDS/mAP results on FocalFormer3D. $^\dagger$ denotes training without self-supervised paradigm in the student model.}
\label{tab:focalformer3d}
\end{table}

\subsection{Ablation Experiments}
In this section, we conduct extensive ablation studies to study the effectiveness of each component of Point-DETR3D. If not specified, the labeled ratio and the backbone are set to 10\% and Voxel-based by default.

\subsubsection{Explicit Positional Query Initialization.}
In Table \ref{tab.ablations.pointencoder}, we compare our explicit positional query initialization strategy with the implicit point encoder proposed in Point-DETR. Simply encoding the $x,y$ coordinates and category information brings a reasonable improvement of 37 mAP over the baseline detector. Then, we replace such implicit initialization with our explicit positional query initialization, and the detection accuracy is enhanced by more than 10.0 mAP compared to the original Point-DETR-style encoding approach. 
We also discover that linking each object query to a unique GT annotation during training alleviates the model's learning challenges, enabling competitive results with just one decoder compared to six.
We conjecture that our one-to-one label assignment strategy stabilizes the training process compared to original Hungarian matching, echoing the observation in \cite{li2022dn}.

\begin{figure}[!t]
\centering
\includegraphics[width=1\linewidth]{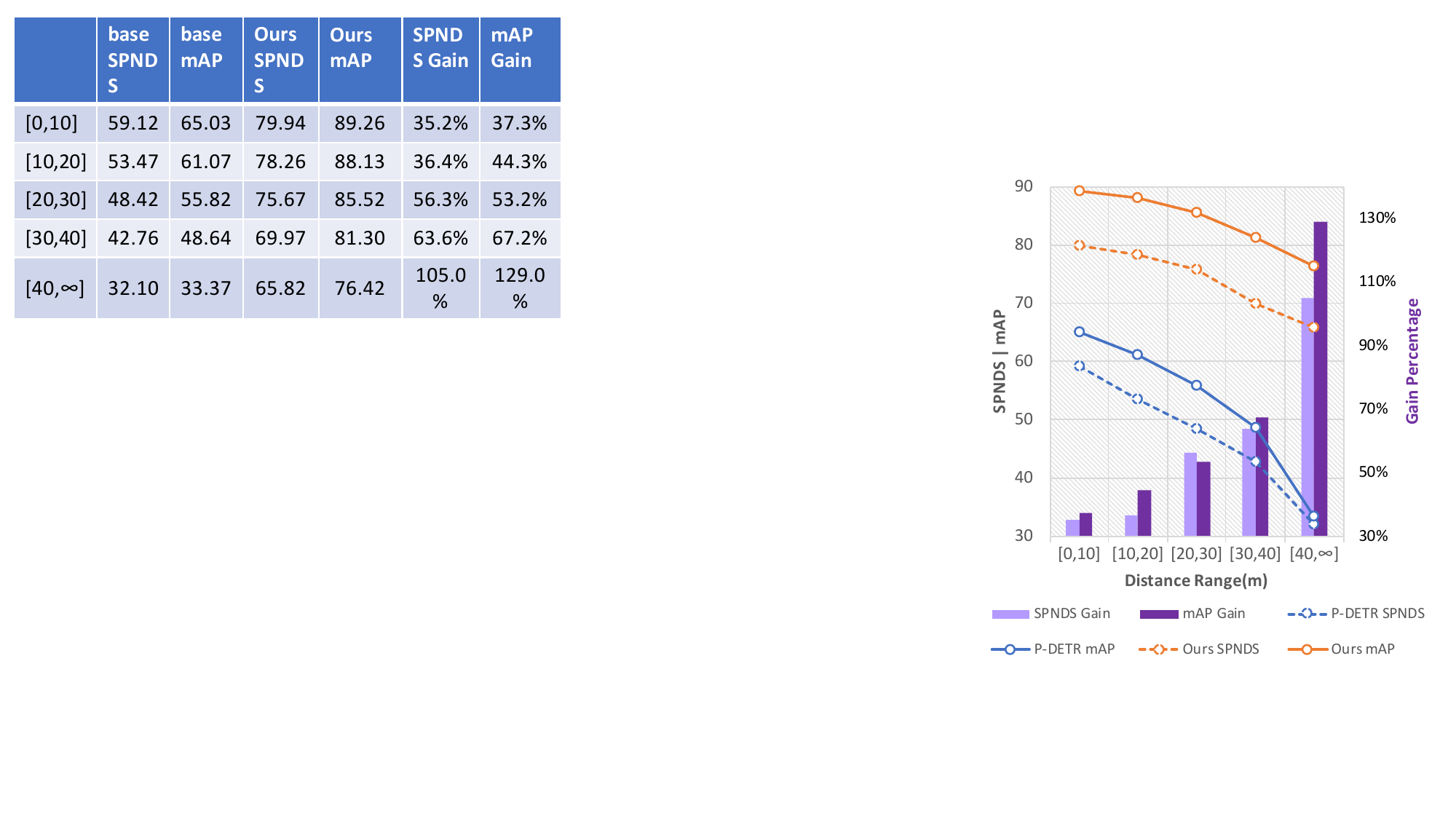}
\vspace{-0.8cm}
\caption{Comparison of Point-DETR3D (Ours) and Point-DETR (P-DETR) in terms of SPNDS, mAP, and Gains in different detection ranges.}
\label{fig:ablation.range_dection}
\end{figure}

\begin{table}[]
\centering
\begin{tabular}{@{}c|cc@{}}
\toprule
Query Initialization & SPNDS & mAP \\ \midrule
 -               & 31.65 & 19.85 \\
Implicit(Point-DETR) & 49.35 & 57.58 \\
Explicit(Ours) & \textbf{58.36(+9.01)} & \textbf{68.67(+11.09)} \\ \bottomrule
\end{tabular}
\caption{The effectiveness of explicit positional query initialization for the teacher model.}
\label{tab.ablations.pointencoder}
\end{table}


\subsubsection{Point-Centric Deformable RoI Cross-Modal Fusion. } 
In this part, we study the effectiveness of our proposed DeformRoICM Fusion module, compared with other point/voxel-based multi-modal fusion strategies, including TransFusion, AutoAlignV2, and MoCa, and report the results in Table \ref{tab:ablation.fusion}. We can find that simply adopting attention-based approaches (TransFusion) yields limited improvement, which is mainly due to the weak positional correspondence during fusion. AutoAlignV2 and MoCa are two point-based fusion strategies, which keep more positional information and achieve 70 SPNDS. 
Furthermore, our proposed DeformRoICM module maximizes the advantage of point priors and yields an impressive 6.2 SPNDS and 7.5 mAP growth by enhancing the perception ability around objects. This is achieved by explicitly restricting the deformable sampling positions to be centered around RoI regions, rather than solely relying on the deformable offsets without any constraints. 
To further investigate the improvements in distant regions, we evaluate the detection performance with different distance intervals and report results in Figure \ref{fig:ablation.range_dection}. As the distance range increases, the detection performance of Point-DETR declines dramatically, which confirms the suppose that LiDAR inputs suffer from extreme sparsity of point clouds. Thanks to the dense visual information brought by imagery features, our approach suffers little performance drop as the distance goes up. Notably, at a distance of more than 40m, Point-DETR3D outperforms Point-DETR, with an increase of 105\% in SPNDS and 129\% in mAP. These results demonstrate that our Point-Centric Teacher can significantly improve the quality of pseudo labels at far distances.


\begin{table}[]
\centering
\begin{tabular}{cc|cc}
\toprule
Fusion Strategy & Venue & SPNDS & mAP \\ \midrule
-  & - & 58.36 & 68.67 \\
 TransFusion & CVPR\citeyear{bai2022transfusion} & 68.54 & 84.03 \\
 AutoAlignV2 & ECCV\citeyear{chen2022autoalignv2} & 70.26 & 79.88 \\
 MoCa  & ICLRW\citeyear{zhang2023exploring} & 69.68 & 78.24 \\
 Ours & - &\textbf{76.46} & \textbf{87.46} \\ \midrule
\end{tabular}
\vspace{-0.2cm}
\caption{The effectiveness of incorporating imagery data(Imagery) and Deformable RoI Fusion Module(D-RoI) for the teacher model. }
\label{tab:ablation.fusion}
\end{table}


\subsubsection{Point-Guided Self-Supervised Learning. } 


We investigate the importance of our point-guided self-supervised paradigm in Table \ref{tab.selfsupervised}. Directly adopting full feature space for contrastive learning can even deteriorate the performance, with a slight drop of 0.2 mAP. We infer the reason that most regions in the BEV feature map are meaningless and noisy. Simply treating all regions equally would amplify the influence of the imprecise pseudo-labels on the student model.
Then, we add the proposed point-guided supervision strategy to dynamically select the informative regions for contrastive learning, which enhances the performance by 1.0 mAP. Such an improvement demonstrates the effectiveness of the utilization of point priors at the student training stage, which is often omitted in previous WSSOD works.

\begin{table}[]
\centering
\begin{tabular}{@{}cc|cc@{}}
\toprule
SSL & Point-guided & SPNDS & mAP \\ \midrule
 &  & 56.72 & 52.04 \\
\checkmark &  & 56.70(-0.02) & 51.85(-0.19) \\
\checkmark & \checkmark & \textbf{57.81(+1.09)} & \textbf{53.03(+0.99)} \\ \bottomrule
\end{tabular}
\vspace{-0.2cm}
\caption{Comparison on different feature-based self-supervised learning(SSL) approaches for the student model.}
\label{tab.selfsupervised}
\end{table}

\vspace{-0.2cm}

\section{Conclusion}
In this work, we introduce Point-DETR3D, a point-centric teacher-student framework for weakly semi-supervised 3D object detection. Our approach capitalizes on spatial point priors, leveraging deformable RoI cross-attention networks to dynamically extract and aggregate RoI-wise features from different modalities.
Additionally, we propose a general point-guided self-supervised learning paradigm for the student model, called Point-Centric Feature-Invariant Learning. This innovative approach can effectively mitigate the impact of label noise, thus enhancing the robustness of the model's representations.
The experiments conducted on the challenging nuScenes benchmark demonstrate that Point-DETR3D outperforms supervised models by over 25 SPNDS and 35 mAP with only 2\% labeled data,
suggesting its potential applicability to various end-to-end LiDAR 3D detectors.
We believe that Point-DETR3D is able to serve as a solid baseline and will inspire further exploration in the weakly semi-supervised 3D object detection area.

\section{Acknowledgments}
This work was supported by the JKW Research Funds under Grant 20-163-14-LZ-001-004-01, and the Anhui Provincial Natural Science Foundation under Grant 2108085UD12. We acknowledge the support of GPU cluster built by MCC Lab of Information Science and Technology Institution, USTC.

\bibliography{aaai24}

\begin{thebibliography}{40}
\providecommand{\natexlab}[1]{#1}

\bibitem[{Bai et~al.(2022{\natexlab{a}})Bai, Hu, Zhu, Huang, Chen, Fu, and
  Tai}]{2022transfusion}
Bai, X.; Hu, Z.; Zhu, X.; Huang, Q.; Chen, Y.; Fu, H.; and Tai, C.~L.
  2022{\natexlab{a}}.
\newblock TransFusion: Robust LiDAR-Camera Fusion for 3D Object Detection with
  Transformers.

\bibitem[{Bai et~al.(2022{\natexlab{b}})Bai, Hu, Zhu, Huang, Chen, Fu, and
  Tai}]{bai2022transfusion}
Bai, X.; Hu, Z.; Zhu, X.; Huang, Q.; Chen, Y.; Fu, H.; and Tai, C.-L.
  2022{\natexlab{b}}.
\newblock Transfusion: Robust lidar-camera fusion for 3d object detection with
  transformers.
\newblock In \emph{Proceedings of the IEEE/CVF conference on computer vision
  and pattern recognition}, 1090--1099.

\bibitem[{Bilen and Vedaldi(2016)}]{bilen2016weakly}
Bilen, H.; and Vedaldi, A. 2016.
\newblock Weakly supervised deep detection networks.
\newblock In \emph{Proceedings of the IEEE conference on computer vision and
  pattern recognition}, 2846--2854.

\bibitem[{Caesar et~al.(2020)Caesar, Bankiti, Lang, Vora, Liong, Xu, Krishnan,
  Pan, Baldan, and Beijbom}]{nuscenes}
Caesar, H.; Bankiti, V.; Lang, A.~H.; Vora, S.; Liong, V.~E.; Xu, Q.; Krishnan,
  A.; Pan, Y.; Baldan, G.; and Beijbom, O. 2020.
\newblock nuscenes: A multimodal dataset for autonomous driving.
\newblock In \emph{Proceedings of the IEEE/CVF conference on computer vision
  and pattern recognition}, 11621--11631.

\bibitem[{Chen et~al.(2021)Chen, Yang, Zhang, Zhang, and Sun}]{chen2021points}
Chen, L.; Yang, T.; Zhang, X.; Zhang, W.; and Sun, J. 2021.
\newblock Points as queries: Weakly semi-supervised object detection by points.
\newblock In \emph{Proceedings of the IEEE/CVF Conference on Computer Vision
  and Pattern Recognition}, 8823--8832.

\bibitem[{Chen et~al.(2020)Chen, Kornblith, Norouzi, and
  Hinton}]{chen2020simple}
Chen, T.; Kornblith, S.; Norouzi, M.; and Hinton, G. 2020.
\newblock A simple framework for contrastive learning of visual
  representations.
\newblock In \emph{International conference on machine learning}, 1597--1607.
  PMLR.

\bibitem[{Chen and He(2021)}]{chen2021exploring}
Chen, X.; and He, K. 2021.
\newblock Exploring simple siamese representation learning.
\newblock In \emph{Proceedings of the IEEE/CVF Conference on Computer Vision
  and Pattern Recognition}, 15750--15758.

\bibitem[{Chen et~al.(2022{\natexlab{a}})Chen, Li, Zhang, Fang, Jiang, and
  Zhao}]{chen2022autoalignv2}
Chen, Z.; Li, Z.; Zhang, S.; Fang, L.; Jiang, Q.; and Zhao, F.
  2022{\natexlab{a}}.
\newblock Autoalignv2: Deformable feature aggregation for dynamic multi-modal
  3d object detection.
\newblock \emph{arXiv preprint arXiv:2207.10316}.

\bibitem[{Chen et~al.(2022{\natexlab{b}})Chen, Li, Zhang, Fang, Jiang, and
  Zhao}]{chen2022graph}
Chen, Z.; Li, Z.; Zhang, S.; Fang, L.; Jiang, Q.; and Zhao, F.
  2022{\natexlab{b}}.
\newblock Graph-DETR3D: rethinking overlapping regions for multi-view 3D object
  detection.
\newblock In \emph{Proceedings of the 30th ACM International Conference on
  Multimedia}, 5999--6008.

\bibitem[{Chen et~al.(2022{\natexlab{c}})Chen, Li, Zhang, Fang, Jiang, Zhao,
  Zhou, and Zhao}]{chen2022autoalign}
Chen, Z.; Li, Z.; Zhang, S.; Fang, L.; Jiang, Q.; Zhao, F.; Zhou, B.; and Zhao,
  H. 2022{\natexlab{c}}.
\newblock Autoalign: Pixel-instance feature aggregation for multi-modal 3d
  object detection.
\newblock \emph{arXiv preprint arXiv:2201.06493}.

\bibitem[{Fan et~al.(2021)Fan, Pang, Zhang, Wang, Zhao, Wang, Wang, and
  Zhang}]{2021Embracing}
Fan, L.; Pang, Z.; Zhang, T.; Wang, Y.~X.; Zhao, H.; Wang, F.; Wang, N.; and
  Zhang, Z. 2021.
\newblock Embracing Single Stride 3D Object Detector with Sparse Transformer.
\newblock \emph{arXiv e-prints}.

\bibitem[{Ge et~al.(2023)Ge, Zhou, Wang, Shen, Wang, and Li}]{ge2023point}
Ge, Y.; Zhou, Q.; Wang, X.; Shen, C.; Wang, Z.; and Li, H. 2023.
\newblock Point-teaching: weakly semi-supervised object detection with point
  annotations.
\newblock In \emph{Proceedings of the AAAI Conference on Artificial
  Intelligence}, volume~37, 667--675.

\bibitem[{Kim et~al.(2023)Kim, Jeong, Han, and Hwang}]{kim2023devil}
Kim, B.; Jeong, J.; Han, D.; and Hwang, S.~J. 2023.
\newblock The Devil is in the Points: Weakly Semi-Supervised Instance
  Segmentation via Point-Guided Mask Representation.
\newblock In \emph{Proceedings of the IEEE/CVF Conference on Computer Vision
  and Pattern Recognition}, 11360--11370.

\bibitem[{Lang et~al.(2019)Lang, Vora, Caesar, Zhou, Yang, and
  Beijbom}]{lang2019pointpillars}
Lang, A.~H.; Vora, S.; Caesar, H.; Zhou, L.; Yang, J.; and Beijbom, O. 2019.
\newblock Pointpillars: Fast encoders for object detection from point clouds.
\newblock In \emph{Proceedings of the IEEE/CVF conference on computer vision
  and pattern recognition}, 12697--12705.

\bibitem[{Lee et~al.(2019)Lee, Kim, Lee, Lee, and Yoon}]{lee2019ficklenet}
Lee, J.; Kim, E.; Lee, S.; Lee, J.; and Yoon, S. 2019.
\newblock Ficklenet: Weakly and semi-supervised semantic image segmentation
  using stochastic inference.
\newblock In \emph{Proceedings of the IEEE/CVF Conference on Computer Vision
  and Pattern Recognition}, 5267--5276.

\bibitem[{Li et~al.(2022)Li, Zhang, Liu, Guo, Ni, and Zhang}]{li2022dn}
Li, F.; Zhang, H.; Liu, S.; Guo, J.; Ni, L.~M.; and Zhang, L. 2022.
\newblock Dn-detr: Accelerate detr training by introducing query denoising.
\newblock In \emph{Proceedings of the IEEE/CVF Conference on Computer Vision
  and Pattern Recognition}, 13619--13627.

\bibitem[{Li, Arnab, and Torr(2018)}]{li2018weaklysemi}
Li, Q.; Arnab, A.; and Torr, P.~H. 2018.
\newblock Weakly-and semi-supervised panoptic segmentation.
\newblock In \emph{Proceedings of the European conference on computer vision
  (ECCV)}, 102--118.

\bibitem[{Liang et~al.(2021)Liang, Jiang, Feng, Chen, Xu, Liang, Zhang, Li, and
  Van~Gool}]{liang2021expcontrast}
Liang, H.; Jiang, C.; Feng, D.; Chen, X.; Xu, H.; Liang, X.; Zhang, W.; Li, Z.;
  and Van~Gool, L. 2021.
\newblock Exploring geometry-aware contrast and clustering harmonization for
  self-supervised 3d object detection.
\newblock In \emph{Proceedings of the IEEE/CVF International Conference on
  Computer Vision}, 3293--3302.

\bibitem[{Liang et~al.(2022)Liang, Xie, Yu, Xia, Lin, Wang, Tang, Wang, and
  Tang}]{liang2022bevfusion}
Liang, T.; Xie, H.; Yu, K.; Xia, Z.; Lin, Z.; Wang, Y.; Tang, T.; Wang, B.; and
  Tang, Z. 2022.
\newblock Bevfusion: A simple and robust lidar-camera fusion framework.
\newblock \emph{Advances in Neural Information Processing Systems}, 35:
  10421--10434.

\bibitem[{Meng et~al.(2020)Meng, Wang, Zhou, Shen, Van~Gool, and
  Dai}]{meng2020weakly}
Meng, Q.; Wang, W.; Zhou, T.; Shen, J.; Van~Gool, L.; and Dai, D. 2020.
\newblock Weakly supervised 3d object detection from lidar point cloud.
\newblock In \emph{Computer Vision--ECCV 2020: 16th European Conference,
  Glasgow, UK, August 23--28, 2020, Proceedings, Part XIII}, 515--531.
  Springer.

\bibitem[{Papandreou et~al.(2015)Papandreou, Chen, Murphy, and
  Yuille}]{papandreou2015weaklysemi}
Papandreou, G.; Chen, L.-C.; Murphy, K.~P.; and Yuille, A.~L. 2015.
\newblock Weakly-and semi-supervised learning of a deep convolutional network
  for semantic image segmentation.
\newblock In \emph{Proceedings of the IEEE international conference on computer
  vision}, 1742--1750.

\bibitem[{Park et~al.(2022)Park, Xu, Zhou, Tomizuka, and
  Zhan}]{park2022detmatch}
Park, J.; Xu, C.; Zhou, Y.; Tomizuka, M.; and Zhan, W. 2022.
\newblock Detmatch: Two teachers are better than one for joint 2d and 3d
  semi-supervised object detection.
\newblock In \emph{Computer Vision--ECCV 2022: 17th European Conference, Tel
  Aviv, Israel, October 23--27, 2022, Proceedings, Part X}, 370--389. Springer.

\bibitem[{Peng et~al.(2022)Peng, Yan, Wu, Yang, He, and Cai}]{peng2022weakm3d}
Peng, L.; Yan, S.; Wu, B.; Yang, Z.; He, X.; and Cai, D. 2022.
\newblock Weakm3d: Towards weakly supervised monocular 3d object detection.
\newblock \emph{arXiv preprint arXiv:2203.08332}.

\bibitem[{Su, Deng, and Fei-Fei(2012)}]{su2012crowdsourcing}
Su, H.; Deng, J.; and Fei-Fei, L. 2012.
\newblock Crowdsourcing annotations for visual object detection.
\newblock In \emph{Workshops at the twenty-sixth AAAI conference on artificial
  intelligence}. Citeseer.

\bibitem[{Tang et~al.(2017)Tang, Wang, Bai, and Liu}]{tang2017multiple}
Tang, P.; Wang, X.; Bai, X.; and Liu, W. 2017.
\newblock Multiple instance detection network with online instance classifier
  refinement.
\newblock In \emph{Proceedings of the IEEE conference on computer vision and
  pattern recognition}, 2843--2851.

\bibitem[{Tarvainen and Valpola(2017)}]{tarvainen2017mean}
Tarvainen, A.; and Valpola, H. 2017.
\newblock Mean teachers are better role models: Weight-averaged consistency
  targets improve semi-supervised deep learning results.
\newblock \emph{Advances in neural information processing systems}, 30.

\bibitem[{Vaswani et~al.(2017)Vaswani, Shazeer, Parmar, Uszkoreit, Jones,
  Gomez, Kaiser, and Polosukhin}]{vaswani2017attention}
Vaswani, A.; Shazeer, N.; Parmar, N.; Uszkoreit, J.; Jones, L.; Gomez, A.~N.;
  Kaiser, {\L}.; and Polosukhin, I. 2017.
\newblock Attention is all you need.
\newblock \emph{Advances in neural information processing systems}, 30.

\bibitem[{Wang et~al.(2021)Wang, Cong, Litany, Gao, and
  Guibas}]{wang20213dioumatch}
Wang, H.; Cong, Y.; Litany, O.; Gao, Y.; and Guibas, L.~J. 2021.
\newblock 3dioumatch: Leveraging iou prediction for semi-supervised 3d object
  detection.
\newblock In \emph{Proceedings of the IEEE/CVF Conference on Computer Vision
  and Pattern Recognition}, 14615--14624.

\bibitem[{Wang et~al.(2022)Wang, Guizilini, Zhang, Wang, Zhao, and
  Solomon}]{wang2022detr3d}
Wang, Y.; Guizilini, V.~C.; Zhang, T.; Wang, Y.; Zhao, H.; and Solomon, J.
  2022.
\newblock Detr3d: 3d object detection from multi-view images via 3d-to-2d
  queries.
\newblock In \emph{Conference on Robot Learning}, 180--191. PMLR.

\bibitem[{Wang and Solomon(2021)}]{wang2021object}
Wang, Y.; and Solomon, J.~M. 2021.
\newblock Object dgcnn: 3d object detection using dynamic graphs.
\newblock \emph{Advances in Neural Information Processing Systems}, 34:
  20745--20758.

\bibitem[{Xie et~al.(2023)Xie, Xu, Rakotosaona, Rim, Tombari, Keutzer,
  Tomizuka, and Zhan}]{xie2023sparsefusion}
Xie, Y.; Xu, C.; Rakotosaona, M.-J.; Rim, P.; Tombari, F.; Keutzer, K.;
  Tomizuka, M.; and Zhan, W. 2023.
\newblock SparseFusion: Fusing Multi-Modal Sparse Representations for
  Multi-Sensor 3D Object Detection.
\newblock \emph{arXiv preprint arXiv:2304.14340}.

\bibitem[{Xu et~al.(2022)Xu, Wang, Zheng, Rao, Zhou, and Lu}]{xu2022back}
Xu, X.; Wang, Y.; Zheng, Y.; Rao, Y.; Zhou, J.; and Lu, J. 2022.
\newblock Back to reality: Weakly-supervised 3d object detection with
  shape-guided label enhancement.
\newblock In \emph{Proceedings of the IEEE/CVF Conference on Computer Vision
  and Pattern Recognition}, 8438--8447.

\bibitem[{Yang, Li, and Dou(2019)}]{yang2019towards}
Yang, K.; Li, D.; and Dou, Y. 2019.
\newblock Towards precise end-to-end weakly supervised object detection
  network.
\newblock In \emph{Proceedings of the IEEE/CVF International Conference on
  Computer Vision}, 8372--8381.

\bibitem[{Yang et~al.(2022)Yang, Chen, Miao, Li, Zhu, and
  Zhang}]{yang2022deepinteraction}
Yang, Z.; Chen, J.; Miao, Z.; Li, W.; Zhu, X.; and Zhang, L. 2022.
\newblock Deepinteraction: 3d object detection via modality interaction.
\newblock \emph{Advances in Neural Information Processing Systems}, 35:
  1992--2005.

\bibitem[{Yin et~al.(2022)Yin, Fang, Zhou, Zhang, Xu, Shen, and
  Wang}]{yin2022semi}
Yin, J.; Fang, J.; Zhou, D.; Zhang, L.; Xu, C.-Z.; Shen, J.; and Wang, W. 2022.
\newblock Semi-supervised 3D object detection with proficient teachers.
\newblock In \emph{Computer Vision--ECCV 2022: 17th European Conference, Tel
  Aviv, Israel, October 23--27, 2022, Proceedings, Part XXXVIII}, 727--743.
  Springer.

\bibitem[{Yin, Zhou, and Krahenbuhl(2021)}]{yin2021center}
Yin, T.; Zhou, X.; and Krahenbuhl, P. 2021.
\newblock Center-based 3d object detection and tracking.
\newblock In \emph{Proceedings of the IEEE/CVF conference on computer vision
  and pattern recognition}, 11784--11793.

\bibitem[{Zhang et~al.(2022)Zhang, Yu, Liu, Wang, Zhou, and
  Chen}]{zhang2022group}
Zhang, S.; Yu, Z.; Liu, L.; Wang, X.; Zhou, A.; and Chen, K. 2022.
\newblock Group R-CNN for weakly semi-supervised object detection with points.
\newblock In \emph{Proceedings of the IEEE/CVF Conference on Computer Vision
  and Pattern Recognition}, 9417--9426.

\bibitem[{Zhang, Wang, and Loy(2023)}]{zhang2023exploring}
Zhang, W.; Wang, Z.; and Loy, C.~C. 2023.
\newblock Improving Data Augmentation for Multi-Modality 3D Object Detection.
\newblock In \emph{International Conference on Learning Representations
  Workshop on Scene Representations for Autonomous Driving}, 1--10.

\bibitem[{Zhao, Chua, and Lee(2020)}]{zhao2020sess}
Zhao, N.; Chua, T.-S.; and Lee, G.~H. 2020.
\newblock Sess: Self-ensembling semi-supervised 3d object detection.
\newblock In \emph{Proceedings of the IEEE/CVF Conference on Computer Vision
  and Pattern Recognition}, 11079--11087.

\bibitem[{Zhou and Tuzel(2018)}]{zhou2018voxelnet}
Zhou, Y.; and Tuzel, O. 2018.
\newblock Voxelnet: End-to-end learning for point cloud based 3d object
  detection.
\newblock In \emph{Proceedings of the IEEE conference on computer vision and
  pattern recognition}, 4490--4499.

\end{thebibliography}

\end{document}